\documentclass[10pt,twocolumn,letterpaper]{article}

\usepackage{wacv}
\usepackage{times}
\usepackage{epsfig}
\usepackage{graphicx}
\usepackage{amsmath}
\usepackage{amssymb}

\usepackage[pagebackref=true,breaklinks=true,letterpaper=true,colorlinks,bookmarks=false]{hyperref}

\wacvfinalcopy 

\def\wacvPaperID{196} 

\ifwacvfinal\fi
\begin{document}
	
\title{Subcategory-aware Convolutional Neural Networks for \\ Object Proposals and Detection}


\author{Yu Xiang$^{1}$, Wongun Choi$^{2}$, Yuanqing Lin$^{3}$, and Silvio Savarese$^{4}$\\
	$^{1}$University of Washington,  $^{2}$NEC Laboratories America, Inc., $^{3}$Baidu, Inc., $^{4}$Stanford University \\
	{\tt\small yuxiang@cs.washington.edu, wongun@nec-labs.com, linyuanqing@baidu.com, ssilvio@stanford.edu}
}

\maketitle
\ifwacvfinal\thispagestyle{empty}\fi

\begin{abstract}
	
In Convolutional Neural Network (CNN)-based object detection methods, region proposal becomes a bottleneck when objects exhibit significant scale variation, occlusion or truncation. In addition, these methods mainly focus on 2D object detection and cannot estimate detailed properties of objects. In this paper, we propose subcategory-aware CNNs for object detection. We introduce a novel region proposal network that uses subcategory information to guide the proposal generating process, and a new detection network for joint detection and subcategory classification. By using subcategories related to object pose, we achieve state-of-the-art performance on both detection and pose estimation on commonly used benchmarks.

\end{abstract}

\vspace{-4mm}
\section{Introduction}

Convolutional Neural Networks (CNNs) have become dominating in solving different recognition problems recently. CNNs are powerful due to their capability in both representation and learning. With millions of weights in the contemporary CNNs, they are able to learn much richer representations from data. In object detection, we have witnessed the performance boost when CNNs \cite{krizhevsky2012imagenet,simonyan2014very} are applied to commonly used benchmarks such as PASCAL VOC \cite{pascal-voc-2012} and ImageNet \cite{ILSVRC15}.

However, there are two main limitations of the state-of-the-art CNN-based object detection methods \cite{girshick2013rich,girshick2015fast,ren2015faster}. First, they rely on region proposal methods \cite{uijlings2013selective,zitnick2014edge,arbelaez2014multiscale} to generate object candidates, which are often based on low-level image features such as superpixels or edges. Although these methods work very well on PASCAL VOC \cite{pascal-voc-2012} and ImageNet \cite{ILSVRC15}, however, when it comes to the KITTI dataset for autonomous driving \cite{geiger2012we} where objects have large scale variation, occlusion and truncation, these region proposal methods perform very poor as observed in our experiments. Recently, the Region Proposal Network (RPN) in \cite{ren2015faster} is able to improve over the traditional region proposal methods. However, it still cannot efficiently handle the scale change of object, occlusion and truncation. Second, the existing CNN-based object detection methods mainly focus on 2D object detection with bounding boxes. As a result, they are not able to estimate detailed information about objects such as 2D segmentation boundary, 3D pose or occlusion relationship between objects, while these information is critical for various applications such as autonomous driving, robotics and augmented reality. 

\begin{figure} [t]\small
	\centering
	\includegraphics[height = 0.35\linewidth,width = 0.9\linewidth]{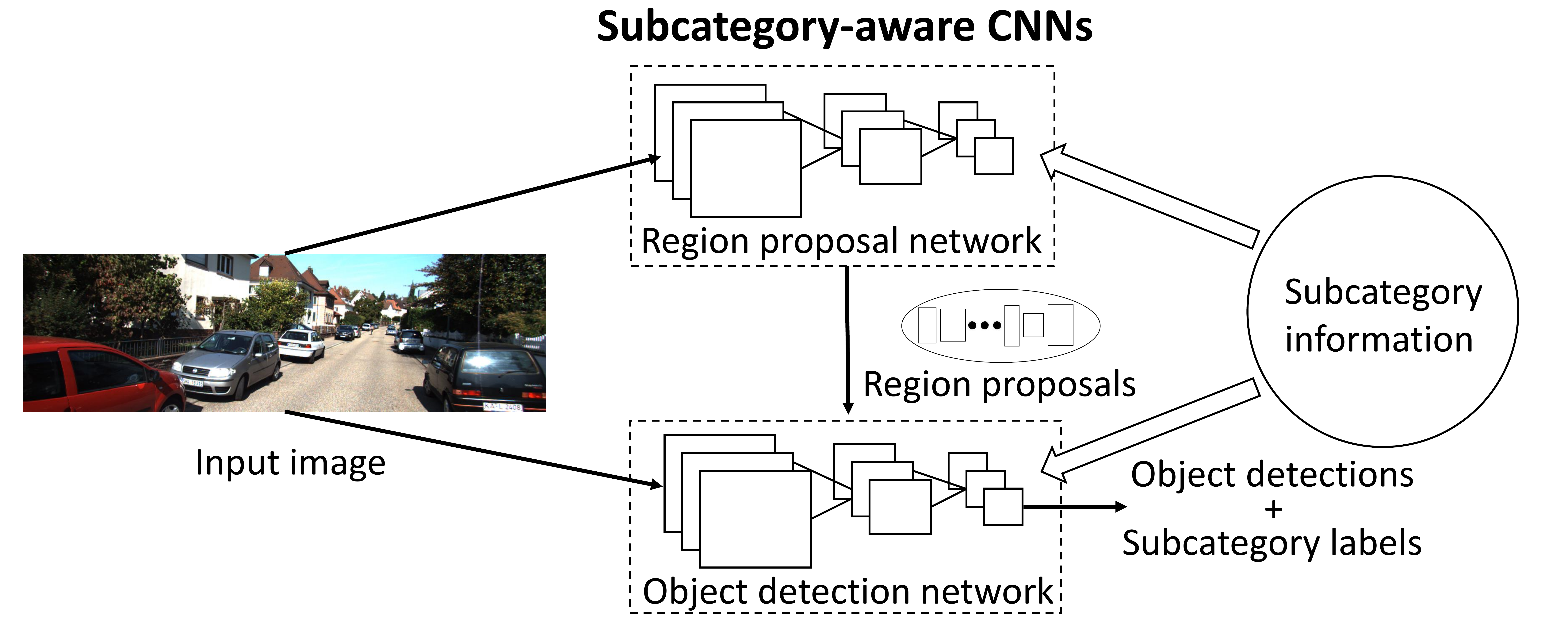}
	\caption{Overview of our object detection framework. By exploiting subcategory information, we propose a new CNN architecture for region proposal and a new object detection network for joint detection and subcategory classification.}
	\label{fig:intro}
	\vspace{-4mm}
\end{figure}

In this work, we explore subcategory information, which is widely used in traditional object detection \cite{felzenszwalb2010object,subcat15,xiang2015data}, to tackle the aforementioned two limitations in CNN-based object detection. For region proposal generation, we introduce a new CNN architecture that uses subcategory detections as object candidates. For detection, we modify the network in Fast R-CNN \cite{girshick2015fast} for joint detection and subcategory classification. Fig. \ref{fig:intro} illustrates our object detection framework. The concept of subcategory is general here. A subcategory can be objects with similar properties or attributes such as 2D appearance, 3D pose or 3D shape. By associating object attributes to subcategories, we are able to estimate these attributes (e.g., 2D segmentation boundary or 3D pose) by conducting subcategory classification.

Specifically, motivated by the traditional detection methods that train a template or a detector for each subcategory, we introduce a \emph{subcategory convolutional (conv) layer} in our Region Proposal Network (RPN), where each filter in the conv layer is trained discriminatively for subcategory detection. The subcategory conv layer outputs heat maps about the presence of certain subcategories at a specific location and scale. Using these heat maps, our RPN is able to output confident subcategory detections as proposals. For classifying region proposals and refining their locations, we introduce a new object detection network by injecting subcategory information into the network proposed in Fast R-CNN \cite{girshick2015fast}. Our detection network is able to perform object detection and subcategory classification jointly. By using 3D Voxel Patterns (3DVPs) \cite{xiang2015data} as subcategories, our method is able to jointly detect the object, estimate its 3D pose, segment its boundary and estimate its occluded or truncated regions. In addition, in both our RPN and our detection CNN, we use image pyramids as input, and we introduce a new \emph{feature extrapolating layer} to efficiently compute conv features in multiple scales. In this way, our method is able to detect objects with large scale variations.

We conduct experiments on the KITTI dataset \cite{geiger2012we}, the PASCAL3D+ dataset \cite{xiang2014beyond} and the PASCAL VOC 2007 dataset \cite{Everingham10}. Comparisons with the state-of-the-art methods on these benchmarks demonstrate the advantages of our subcategory-aware CNNs for object recognition.

\section{Related Work}

\noindent \textbf{Subcategory in Object Detection.} Subcategory has been widely utilized to facilitate object detection, and different methods of discovering object subcategories have been proposed. In DPM \cite{felzenszwalb2010object}, subcategories are discovered by clustering objects according to the aspect ratio of their bounding boxes. \cite{gu2010discriminative} performs clustering according to the viewpoint of the object to discover subcategories. Visual subcategories are constructed by clustering in the appearance space of object \cite{divvala2012important,subcat15,chen2014enriching,divvalalearning}. 3DVP \cite{xiang2015data} performs clustering in the 3D voxel space according to the visibility of the voxels. Unlike previous works, we utilize subcategory to improve CNN-based detection, and our framework is general to employ different types of object subcategories.

\noindent \textbf{CNN-based Object Detection.} We can categorize the state-of-the-art CNN-based object detection methods into two classes: one-stage detection and two-stage detection. In one-stage detection, such as the Overfeat \cite{sermanet2013overfeat} framework, a CNN directly processes an input image, and outputs object detections. In two-stage detection, such as R-CNNs \cite{girshick2013rich,girshick2015fast,ren2015faster}, region proposals are first generated from an input image, where different region proposal methods can be employed \cite{uijlings2013selective,zitnick2014edge,arbelaez2014multiscale}. Then these region proposals are fed into a CNN for classification and location refinement. It is debatable which detection paradigm is better. We adopt the two-stage detection framework in this work, and consider the region proposal process to be the coarse detection step in coarse-to-fine detection \cite{viola2004robust}. We propose a novel RPN motivated by \cite{ren2015faster} and demonstrate its advantages.

\section{Subcategory-aware RPN}

Ideally, we want to have a region proposal approach that can cover objects in an input image with as few proposals as possible. Since objects in images appear at different locations and scales, region proposal itself is a challenging problem. Recently, \cite{ren2015faster} proposed to tackle the region proposal problem with CNNs, demonstrating the advantages of using CNNs over traditional approaches for region proposal. In this section, we describe our subcategory-aware Region Proposal Network (RPN).

\subsection{Network Architecture}

\begin{figure*} \small
	\centering
	\includegraphics[height = 0.18\linewidth,width = \linewidth]{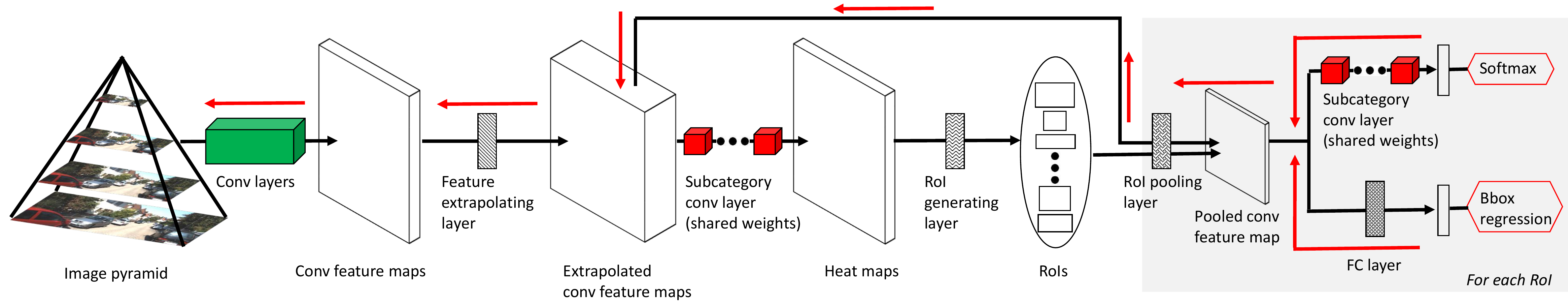}
	\caption{Architecture of our region proposal network. Red arrows indicate the route of derivatives in back-propagation training.}
	\label{fig:rpn}
	\vspace{-4mm}
\end{figure*}

We introduce a novel network architecture for generating object proposals from images. The architecture is inspired by the traditional sliding-window-based object detectors, such as the Aggregated Channel Feature (ACF) detector \cite{DollarPAMI14pyramids} and the Deformable Part Model (DPM) \cite{felzenszwalb2010object}. Fig. \ref{fig:rpn} illustrates the architecture of our region proposal network. i) To handle different scales of objects, we input into our RPN an image pyramid. This pyramid is processed by several convolutional (conv) and max pooling layers to extract the conv feature maps, with one conv feature map for each scale. ii) In order to speed up the computation of conv features on image pyramids, we introduce the \emph{feature extrapolating layer}, which generates feature maps for scales that are not covered by the image pyramid via extrapolation. iii) After computing the extrapolated conv feature maps, we specifically design a conv layer for object subcategory detection, where each filter in the conv layer corresponds to an object subcategory. We train these filters to make sure they fire on correct locations and scales of objects in the corresponding subcategories during the network training. The \emph{subcategory conv layer} outputs a heat map for each scale, where each value in the heat map indicates the confidence of an object in the corresponding location, scale and subcategory. v) Using the subcategory heat maps, we design a \emph{RoI generating layer} that generates object candidates (RoIs) by thresholding the heat maps. vi) The RoIs are used in a \emph{RoI pooling layer} \cite{girshick2015fast} to pool conv features from the extrapolated conv feature maps. vii) Finally, our RPN terminates at two sibling layers: one that outputs softmax probability estimates over object subcategories, and the other layer that refines the RoI location with a bounding box regressor.

\subsection{Feature Extrapolating Layer}

In our RPN, we use fixed-size conv filters in the subcategory conv layer to localize objects (e.g., $5\times5$ conv filters). In order to handle different scales of objects, we resort to image pyramids. An image pyramid consists of images with different resolutions obtained by rescaling the original image according to different sampled scales. After constructing the image pyramid for an input image, multi-resolution conv feature maps can be computed by applying several conv layers and max pooling layers to each image in the pyramid (Fig. \ref{fig:rpn}). If we perform convolution on every scale explicitly, it is computationally expensive, especially when a finely-sampled image pyramid is needed as in the region proposal process. In \cite{DollarPAMI14pyramids}, Doll\'{a}r et al. demonstrate that multi-resolution image features can be approximated by extrapolation from nearby scales rather than being computed explicitly. Inspired by their work, we introduce a feature extrapolating layer to accelerate the computation of conv features on an image pyramid.

Specifically, a feature extrapolating layer takes as input $N$ feature maps that are supplied by the last conv layer for feature extraction, where $N$ equals to the number of scales in the input image pyramid. Each feature map is a multi-dimensional array of size $H \times W \times C$, with $H$ rows, $W$ columns, and $C$ channels. The width and height of the feature map corresponds to the largest scale in the image pyramid, where images in smaller scales are padded with zeros in order to generate feature maps with the same size. The feature extrapolating layer constructs feature maps at intermediate scales by extrapolating features from the nearest scales among the $N$ scales using bilinear interpolation. Suppose we add $M$ intermediate scales between every $i$th scale and $(i+1)$th scale, $i=1,\ldots,N-1$. The output of the feature extrapolating layer is $N'=(N-1)M + N$ feature maps, each with size $H \times W \times C$. Since extrapolating a multi-dimensional array is much faster than computing a conv feature map explicitly, the feature extrapolating layer speeds up the feature computation with less memory.

\subsection{Subcategory Conv Layer}

After computing the conv feature maps, we design a subcategory conv layer for subcategory detection. Motivated by the traditional object detection methods that train a classifier or a template for each subcategory \cite{felzenszwalb2010object,malisiewicz2011ensemble,xiang2015data}, we train a conv filter in the subcategory conv layer to detect a specific subcategory. Suppose there are $K$ subcategories to be considered. Then, the subcategory conv layer consists of $K+1$ conv filters with one additional conv filter for a special ``background'' category. For multi-class detection (e.g., car, pedestrian, cyclist, etc.), the $K$ subcategories are the aggregation of all the subcategories from all the classes. These conv filters operate on the extrapolated conv feature maps and output heat maps that indicate the confidences of the presence of objects in the input image. We use fixed-size conv filters in this layer (e.g., $5 \times 5 \times C$ conv filters), which are trained to fire on specific scales in the feature pyramid. Sec. \ref{sec:rpn_loss} explains how we back-propagate errors from the loss layer to train these subcategory conv filters.

\begin{figure*} \small 
	\centering
	\includegraphics[height = 0.2\linewidth,width = \linewidth]{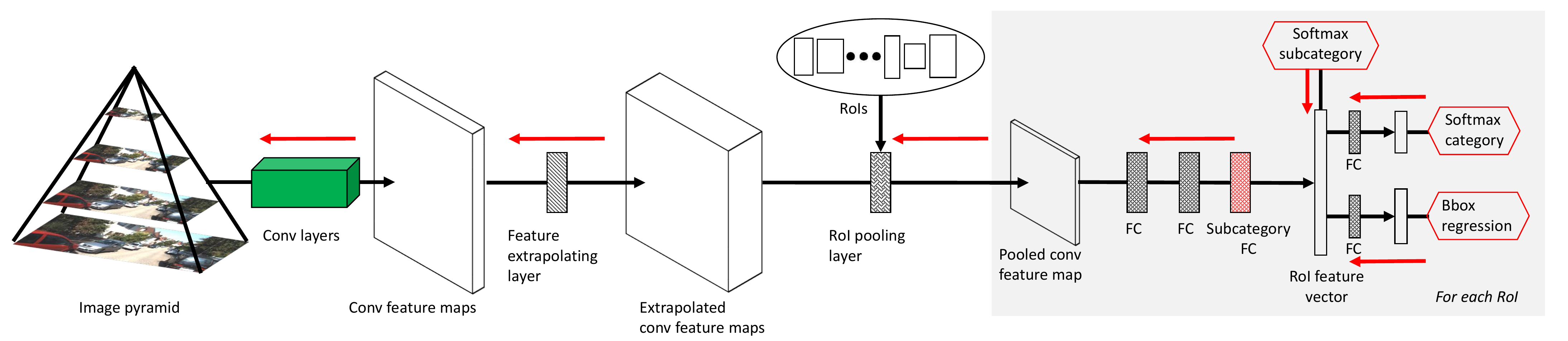}
	\caption{Architecture of our object detection network. Red arrows indicate the route of derivatives in back-propagation training.}
	\label{fig:rcnn}
	\vspace{-4mm}
\end{figure*}

\subsection{RoI Generating Layer}

The RoI generating layer takes as input $N'$ heat maps and outputs a set of region proposals (RoIs), where $N'$ is the number of scales in the feature pyramid after extrapolation. Each heat map is a multi-dimensional array of size $H \times W \times K$ for $K$ subcategories (i.e., for RoI generating, we ignore the ``background'' channel in the heat map). The RoI generating layer first converts each heat map into a $H \times W$ 2D array by performing max operation over the channels for subcategory. Then, it thresholds the 2D heat map to generate RoIs. In this way, we measure the objectness of a region by aggregating information from subcategories. Different generating strategies are used in testing and training.

In testing, each location $(x,y)$ in a heat map with a score larger than a predefined threshold is used to generate RoIs. First, a canonical bounding box is centered on $(x,y)$. The width and height of the box are the same as those of the conv filters (e.g., $5 \times 5 $) in the subcategory conv layer, which have an aspect ratio one. Second, a number of boxes centered on $(x,y)$ with the same areas as the canonical box (e.g., $25$) but with different aspect ratios are generated. Finally, the RoI generating layer rescales the generated boxes according to the scale of the heat map, so as to cover objects in different scales and aspect ratios.

In training, the RoI generating layer outputs hard positive RoIs and hard negative RoIs for training the subcategory conv filters, given a budget on batch size in stochastic gradient descent. First, we use the same procedure as described in testing to generate a number of bounding boxes for each location in each heat map. Second, according to the ground truth bounding boxes of objects in a training image, we compute the intersection over union (IoU) overlap between the generated boxes and the ground truth boxes. Bounding boxes with IoU overlap larger/smaller than some threshold (e.g., 0.5) are considered to be positive/negative. Finally, given the number of RoIs to be generated for each training image $R$ (i.e., batch size divided by the number of images in a batch), the RoI generating layer outputs $R \times \alpha$ hard positives (i.e., $R \times \alpha$ positive bounding boxes with lowest scores in the heat maps) and $R \times (1-\alpha)$ hard negatives (i.e., $R \times (1-\alpha)$ negative bounding boxes with highest scores in the heat maps), where $\alpha \in (0,1)$ is the percentage of positive examples.

\subsection{Network Training} \label{sec:rpn_loss}

After generating RoIs, we apply the RoI pooling layer proposed in \cite{girshick2015fast} to pool conv features for each RoI. Then the pooled conv features are used for two tasks: subcategory classification and bounding box regression. As illustrated in Fig. \ref{fig:rpn}, our RPN has two sibling output layers. The first layer outputs a discrete probability distribution $p = (p_0, \ldots, p_K)$, over $K+1$ subcategories, which is computed by applying a softmax function over the $K+1$ outputs of the subcategory conv layer. The second layer outputs bounding box regression offsets $t^{k'} = (t_x^{k'}, t_y^{k'}, t_w^{k'}, t_h^{k'}), {k'} = 0, 1, \ldots, K'$ for $K'$ object classes ($K' \ll K$). We parameterize $t^{k'}$ as in \cite{girshick2013rich}, which specifies a scale-invariant translation and log-space width/height shift relative to a RoI.

We employ a multi-task loss as in \cite{girshick2015fast} to train our RPN for subcategory classification and bounding box regression:
\begin{equation}
	L(p,k^*,k'^{*},t,t^*) = L_{\text{subcls}}(p,k^*) + \lambda [k'^{*} \geq 1] L_{\text{loc}}(t,t^*),
\end{equation}
where $k^*$ and $k'^{*}$ are the truth subcategory label and the true class label respectively, $L_{\text{subcls}}(p,k^*) = -\log p_{k^*}$ is the standard cross-entropy loss, $t^{*} = (t_x^{*}, t_y^{*}, t_w^{*}, t_h^{*})$ is the true bounding box regression targets for class $k'^{*}$, and $t = (t_x, t_y, t_w, t_h)$ is the prediction for class $k'^{*}$. We use the smoothed $L_1$ loss defined in \cite{girshick2015fast} for the bounding box regression loss $L_{\text{loc}}(t,t^*)$. The indicator function $[k'^{*} \geq 1]$ indicates that bounding box regression is ignored if the RoI is background (i.e., $k'^* = 0$). $\lambda$ is a predefined weight to balance the two losses.

In training, derivatives from the loss function are back-propagated (see red arrows in Fig. \ref{fig:rpn}). The two subcategory conv layers in our RPN share their weights. These weights/conv filters are updated according to the derivatives from the softmax loss function for subcategory classification, so we are able to train these filters for subcategory detection. There is no derivative flow in computing heat maps using the subcategory conv layer and in the RoI generating layer. Finally, our RPN generates confident subcategory detections as region proposals.



\section{Subcategory-aware Detection Network}
\label{sec:det_net}

After the region proposal process, CNNs are utilized to classify these proposals and refine their locations \cite{girshick2013rich,girshick2015fast,ren2015faster}. Since region proposal significantly reduces the search space, more powerful CNNs can be used in the detection step, which usually contain several fully connected layers with high dimensions. In this section, we introduce our subcategory-aware object detection network for joint detection and subcategory classification.

\subsection{Network Architecture}

Fig. \ref{fig:rcnn} illustrates the architecture of our detection network. The network is constructed based on the Fast R-CNN detection network \cite{girshick2015fast} with a number of improvements. i) We use image pyramids to handle the scale variation of objects. After the last conv layer for feature extraction, we add the feature extrapolating layer to increase the number of scales in the conv feature pyramid. ii) Given the region proposals generated from our RPN, we employ a RoI pooling layer to pool conv features for each RoI. Each RoI is mapped to a scale in the conv feature pyramid such that smaller RoIs pool features from larger scales. iii) The pooled conv features are fed into three fully connected (FC) layers, where the last FC layer is designed for subcategory classification. For $K$ subcategories, the ``subcategory FC'' layer outputs a $K+1$ dimensional vector with one additional dimension for the background class. We consider the output, named \emph{RoI feature vector}, to be an embedding in the subcategory space. iv) Finally, the network terminates at three output layers. The first output layer applies a softmax function directly on the output of the ``subcategory FC'' layer for subcategory classification. The other two output layers operate on the RoI feature vector and apply FC layers for object class classification and bounding box regression. 

\subsection{Network Training}

We train our object detection network with a multi-task loss for joint object class classification, subcategory classification and bounding box regression:
\begin{align}
	& L(p,k^*,p',k'^{*},t,t^*) = \\
	& L_{\text{subcls}}(p,k^*) + \lambda_1 L_{\text{cls}}(p',k'^*) + \lambda_2 [k'^{*} \geq 1] L_{\text{loc}}(t,t^*), \nonumber
\end{align}
where $p = (p_0, \ldots, p_K)$ is a probability distribution over $K+1$ subcategories, $p' = (p'_0, \ldots, p'_{K'})$ is a probability distribution over $K'+1$ object classes, $k^*$ and $k'^{*}$ are the truth subcategory label and the true class label respectively, $t$ and $t^{*}$ are the predicted vector and the true vector for bounding box regression respectively, and $\lambda_1$ and $\lambda_2$ are predefined weights to balance the losses of different tasks. $L_{\text{subcls}}(p,k^*) = -\log p_{k^*}$ and $L_{\text{cls}}(p',k'^*) = -\log p'_{k'^*}$ are the standard cross-entropy loss, and $L_{\text{loc}}(t,t^*)$ is the smoothed $L_1$ loss as in our RPN. In back-propagation training, derivatives for the multi-task loss are back-propagated to the previous layers. Red arrows in Fig. \ref{fig:rcnn} indicate the route of the derivative flow.

\section{Experiments} \label{sec:exp}

\subsection{Experimental Settings}
	
\noindent \textbf{Datasets.} We evaluate our object detection framework on the KITTI detection benchmark \cite{geiger2012we}, the PASCAL3D+ dataset \cite{xiang2014beyond} and the PASCAL VOC 2007 dataset \cite{Everingham10}. i) The KITTI dataset consists of video frames from autonomous driving scenes, with 7,481 images for training and 7,518 images for testing. Car, pedestrian and cyclist are evaluated for object detection. Since the ground truth annotations of the KITTI test set are not released, we split the KITTI training images into a train set and a validation set for analyses as in \cite{xiang2015data}. ii) The PASCAL3D+ dataset augments 12 rigid categories in the PASCAL VOC 2012 \cite{pascal-voc-2012} with 3D annotations. Each object in the 12 categories is registered with a 3D CAD model. The \emph{train} set of PASCAL VOC 2012 is used for training (5,717 images), while the \emph{val} set is used for testing (5,823 images). iii) The PASCAL VOC 2007 dataset \cite{Everingham10} contains 5,011 training images and 4,952 testing images on 20 categories.




\noindent \textbf{Evaluation Metrics.} On KITTI, we evaluate our detection framework at three levels of difficulty as suggested by \cite{KITTIObject}, i.e., easy, moderate and hard, where the difficulty is measured by the minimal scale of object to be considered and the occlusion and truncation of the object. Average Precision (AP) \cite{pascal-voc-2012} is used to measure the detection performance, where 70\%, 50\%, and 50\% overlap thresholds are adopted by the KITTI benchmark for car, pedestrian and cyclist respectively. To evaluate joint detection and orientation estimation on KITTI, \cite{geiger2012we} introduces Average Orientation Similarity (AOS), which evaluates the orientation similarity between detections and ground truths at different detection recalls. \cite{xiang2015data} introduces Average Segmentation Accuracy (ASA) for joint detection and segmentation, and Average Location Precision (ALP) for joint detection and 3D location similar to AOS. We also use these metrics here. On PASCAL3D+ and PASCAL VOC 2007, the standard AP with 50\% overlap ratio is adopted to evaluate object detection. For joint detection and pose estimation, we use the Average Viewpoint Precision (AVP) suggested by \cite{xiang2014beyond}, where a detection is considered to be a true positive if its location and viewpoint are both correct.

\noindent \textbf{Subcategories.} We experiment with both 2D subcategories and 3D subcategories. For 2D subcategories, we cluster objects using 2D image features (i.e., aggregated channel features from \cite{DollarPAMI14pyramids}). Only bounding box annotations are needed for 2D subcategories. When additional annotations are available, we can obtain 3D subcategories. We adopt the 3D Voxel Pattern (3DVP) representation \cite{xiang2015data} for rigid objects (i.e., car in KITTI and the 12 categories in PASCAL3D+), which jointly models object pose, occlusion and truncation in the clustering process. Each 3DVP is considered to be a subcategory. For pedestrian and cyclist in KITTI, we perform clustering according to the object orientation, and each cluster is considered to be a subcategory. In this way, by subcategory classification, we can transfer the meta data carried by 3DVPs (3D pose, segmentation boundary and occluded regions) to the detected object. 

For validation on KITTI (3,682 images for training, 3,799 images for testing), we use 173 subcategories (125 3DVPs for car, 24 poses for pedestrian and cyclist each), while for testing on KITTI (7,481 images for training, 7,518 images for testing), we use 275 subcategories (227 3DVPs for car, 24 poses for pedestrian and cyclist each). 3DVPs are discovered with affinity propagation clustering \cite{frey2007clustering}, which automatically discovers the number of clusters from the data.  For PASCAL3D+, 337 3DVPs are discovered among the 12 categories. For PASCAL VOC 2007, we use 240 2D subcategories, with 12 for each class. Correspondingly, the output number of the subcategory conv layer in our RPN and that of the subcategory FC layer in our detection network equal to the number of subcategory plus one.

\noindent \textbf{Region Proposal Network Hyper-parameters.} In our RPN, we use 5 scales for KITTI in the input image pyramid $(0.25, 0.5, 1.0, 2.0, 3.0)$ and 4 scales for PASCAL $(0.25, 0.5, 1.0, 2.0)$ (both PASCAL3D+ and PASCAL VOC 2007), where each number indicates the rescaling factor with respect to the original image size. Objects in PASCAL have smaller scale variation compared to objects in KITTI. Adding larger scales for PASCAL only results in marginal improvement but significantly increases the computation. The feature extrapolating layer extrapolates 4 scales with equal intervals between every two input scales, so the final conv feature pyramid has 21 scales for KITTI and 16 scales for PASCAL. In the RoI generating layer, each location in a heat map generates 7 boxes with 7 different aspect ratios $(3.0, 2.0, 1.5, 1.0, 0.75, 0.5, 0.25)$ for KITTI and 5 aspect ratios $(3.0, 2.0, 1.0, 0.5, 0.25)$ for PASCAL, where each number indicates the ratio between the height and the width of the bounding box. In training the RPN, each SGD mini-batch is constructed from a single image, chosen uniformly at random. A mini-batch has size 128, with 64 positive RoIs and 64 negative RoIs, where the IoU threshold is 70\% for both KITTI and PASCAL.

\noindent \textbf{Detection Network Hyper-parameters.} In our detection network, we use 4 scales in the input image pyramid $(1.0, 2.0, 3.0, 4.0)$ for KITTI and 2 scales $(1.0, 2.0)$ for PASCAL, both with 4 scales extrapolated between every two scales. Each SGD mini-batch is constructed from 2 images. A mini-batch has size 128, with 64 RoIs from each image. 25\% of the RoIs are positive, where the IoU threshold is 70\% for car in KITTI, and 50\% for the other categories. The same SGD hyper-parameters are used as in \cite{girshick2015fast} for region proposal and detection.

\noindent \textbf{Fine-tuning Pre-trained Networks.} Our framework is implemented in Caffe \cite{jia2014caffe}. We initialize the conv layers for feature extraction in both networks and the two FC layers before subcategory FC layer in the detection network with pre-trained networks on ImageNet \cite{ILSVRC15}. On KITTI, we experiment with the AlexNet \cite{krizhevsky2012imagenet}, the VGG16 network \cite{simonyan2014very} and the GoogleNet \cite{googlenet}. On PASCAL, we fine-tune the VGG16 network \cite{simonyan2014very}. 

\subsection{Analysis on KITTI Validation Set}

\begin{table} 
	{\tiny
		\centering{
			\begin{tabular}{|l||c|c|c|}
				\hline Methods  & Easy  & Moderate & Hard \\
				\hline \multicolumn{4}{|c|}{Car} \\
				\hline Selective Search \cite{uijlings2013selective} & 58.17  & 42.12  & 37.62 \\
				\hline Edge Boxes \cite{zitnick2014edge}  & 81.40 & 61.84 & 55.68 \\			
				\hline RPN \cite{ren2015faster}  & 98.84 & \textbf{97.37} & \textbf{95.31}  \\
				\hline \textbf{Ours}  & \textbf{99.27} & 96.28 & 93.14 \\
				\hline \multicolumn{4}{|c|}{Pedestrian} \\
				\hline Selective Search \cite{uijlings2013selective} & 68.95  & 57.65  & 52.57  \\
				\hline Edge Boxes \cite{zitnick2014edge}  & 86.15 & 71.88 & 65.39 \\			
				\hline RPN \cite{ren2015faster}  & 98.88 & 91.69 & 88.64  \\
				\hline \textbf{Ours}  & \textbf{99.44} & \textbf{93.46} & \textbf{91.02}  \\
				\hline \multicolumn{4}{|c|}{Cyclist} \\
				\hline Selective Search \cite{uijlings2013selective} & 57.05  & 49.59  & 49.44  \\
				\hline Edge Boxes \cite{zitnick2014edge}  & 56.11 & 46.52 & 45.72 \\			
				\hline RPN \cite{ren2015faster}  & 96.55 & 91.80 & 89.41  \\
				\hline \textbf{Ours}  & \textbf{99.67} & \textbf{93.03} & \textbf{91.64}  \\								
				\hline
			\end{tabular}
			\caption{Region proposal performance in terms of recall on the KITTI validation set.}
			\label{table:proposal}
			\vspace{-4mm}
		}
	}
\end{table}

\noindent \textbf{Region Proposal Evalutaion on Recall.} We evaluate the detection recall of our RPN and compare it with the state-of-the-art methods in Table \ref{table:proposal} on the KITTI validation set. For each image, we use 2k proposals for all the methods. First, two popular methods that work well on PASCAL VOC \cite{pascal-voc-2012}, Selective Search \cite{uijlings2013selective} and Edge Boxes \cite{zitnick2014edge}, do not perform well on KITTI, mainly because objects in KITTI exhibit more significant scale variation, occlusion and truncation. It is challenging for a bottom-up proposal method to achieve high recall under a small budget (i.e, 2k boxes per image). Second, the RPN in Faster R-CNN \cite{ren2015faster} performs much better than Selective Search and Edge Boxes, which demonstrates the ability of discriminatively trained CNNs for region proposal. But we have to increase its parameter setting from 3 scales and 3 aspect ratios in \cite{ren2015faster} to 10 scales and 7 aspect ratios in order to make it work on KITTI. Finally, our RPN performs on par with Faster R-CNN on car, and outperforms it on pedestrian and cyclist using the same number of proposals per image. Our new architecture can better handle scale variation using image pyramid. It also benefits from data mining hard training examples in our RoI generating layer.

\begin{table}
	{\tiny
		\centering{
			\begin{tabular}{|@{\hspace{0.25em}}l@{\hspace{0.25em}}||c@{\hspace{0.5em}}|@{\hspace{0.25em}}c@{\hspace{0.25em}}|@{\hspace{0.5em}}c||c@{\hspace{0.5em}}|@{\hspace{0.25em}}c@{\hspace{0.25em}}|@{\hspace{0.5em}}c@{\hspace{0.5em}}|}
				\hline & \multicolumn{3}{|c||}{Object Detection (AP) } & \multicolumn{3}{|c|}{Orientation (AOS)}\\
				\hline Methods  & Easy  & Moderate & Hard & Easy  & Moderate & Hard \\
				\hline \multicolumn{7}{|c|}{Car} \\
				\hline RPN \cite{ren2015faster}+Our det. net (unshared)  & 89.29 & 82.58 & 70.12 & 87.70 & 80.47 & 67.83 \\
				\hline RPN \cite{ren2015faster}+Our det. net (shared)  & 87.67 & 82.21 & 70.10 & 86.58 & 80.27 & 67.90 \\
				\hline \textbf{Ours} (unshared)  & \textbf{95.77} & \textbf{86.64} & \textbf{74.07} & \textbf{94.55} & \textbf{85.03} & \textbf{72.21} \\		
				\hline \multicolumn{7}{|c|}{Pedestrian} \\
				\hline RPN \cite{ren2015faster}+Our det. net (unshared)  & 83.07 & 69.32 & 63.46 & 71.43 & 58.67 & 53.58 \\
				\hline RPN \cite{ren2015faster}+Our det. net (shared)  & 82.73 & 68.28 & 62.30 & 70.31 & 56.94 & 51.87 \\
				\hline \textbf{Ours} (unshared)  & \textbf{86.43} & \textbf{69.95} & \textbf{64.03} & \textbf{73.91} & \textbf{58.91} & \textbf{53.79} \\	
				\hline \multicolumn{7}{|c|}{Cyclist} \\
				\hline RPN \cite{ren2015faster}+Our det. net (unshared)  & 69.23 & 54.83 & 51.41 & 61.25 & 46.44 & 43.07 \\
				\hline RPN \cite{ren2015faster}+Our det. net (shared)  & 71.24 & 56.69 & 52.91 & 63.21 & 48.68 & 45.16 \\
				\hline \textbf{Ours} (unshared)  & \textbf{74.92} & \textbf{59.13} & \textbf{55.03} & \textbf{65.79} & \textbf{50.46} & \textbf{46.57} \\													
				\hline
			\end{tabular}
			\caption{AP/AOS comparison using different region proposals but the same detection network on the KITTI validation set.}
			\label{table:det_val}
			\vspace{-4mm}
		}
	}
\end{table}

\noindent \textbf{Region Proposal Evalutaion on Detection and Oritentaion Estimation.} Detection recall measures the coverage of region proposals, which cannot demonstrate the quality of the region proposals for detection. In this experiment, we directly measure the detection and orientation estimation performance using different region proposals. Table \ref{table:det_val} presents the detection and orientation estimation results using RPN in Faster R-CNN \cite{ren2015faster} and the RPN we propose, while keeping the detection network the same as described in Sec. \ref{sec:det_net}. We compare our RPN with two variations of the RPN in Faster R-CNN. For the first model, the RPN and the detection network are trained independently to each other (``unshared''). For the second model, the RPN and the detection network share their conv layers for feature extraction in order to save computation on convolution (``shared''). The sharing is achieved by the four-step alternating optimization training algorithm described in \cite{ren2015faster}. By comparing the two models in Table \ref{table:det_val}, we find that sharing conv layers hurts the performance on car and pedestrian, but improves the performance on cyclist. Car and pedestrian have much more training examples available than cyclist. With enough training data, the RPN and the detection network trained independently can develop conv features suitable for its own task. In this case, shared conv features degrade the performance. However, when the training data is insufficient, sharing conv features can help.

\begin{table}
	{\tiny
		\centering{
			\begin{tabular}{|@{\hspace{0.25em}}l@{\hspace{0.25em}}||c@{\hspace{0.5em}}|@{\hspace{0.25em}}c@{\hspace{0.25em}}|@{\hspace{0.5em}}c||c@{\hspace{0.5em}}|@{\hspace{0.25em}}c@{\hspace{0.25em}}|@{\hspace{0.5em}}c@{\hspace{0.5em}}|}
				\hline & \multicolumn{3}{|c||}{Object Detection (AP) } & \multicolumn{3}{|c|}{Orientation (AOS)}\\
				\hline Methods  & Easy  & Moderate & Hard & Easy  & Moderate & Hard \\
				\hline \multicolumn{7}{|c|}{Car} \\
				\hline RPN \cite{ren2015faster} + \cite{girshick2015fast}  & 82.91 & 77.83 & 66.25 & N/A & N/A & N/A \\				
				\hline Our RPN + \cite{girshick2015fast}  & 95.14 & 85.20 & 72.12 & N/A & N/A & N/A \\
				\hline
				\hline \textbf{Ours} w/o Pose  & 94.66 & 84.94 & 72.43 & N/A & N/A & N/A \\				
				\hline \textbf{Ours} w/o Extra  & 95.51 & 86.29 & 73.68 & 94.26 & 84.69 & 71.80 \\
				\hline \textbf{Ours} Full  & \textbf{95.77} & \textbf{86.64} & \textbf{74.07} & \textbf{94.55} & \textbf{85.03} & \textbf{72.21} \\		
				\hline \multicolumn{7}{|c|}{Pedestrian} \\
				\hline RPN \cite{ren2015faster} + \cite{girshick2015fast}  & 83.31 & 68.39 & 62.56 & N/A & N/A & N/A \\				
				\hline Our RPN + \cite{girshick2015fast}  & 85.96 & 68.55 & 62.55 & N/A & N/A & N/A \\
				\hline
				\hline \textbf{Ours} w/o Pose  & 83.22 & 67.61 & 62.03 & N/A & N/A & N/A \\
				\hline \textbf{Ours} w/o Extra & 84.86 & 68.87 & 63.09 & \textbf{74.05} & \textbf{59.06} & \textbf{54.05} \\
				\hline \textbf{Ours} Full  & \textbf{86.43} & \textbf{69.95} & \textbf{64.03} & 73.91 & 58.91 & 53.79 \\	
				\hline \multicolumn{7}{|c|}{Cyclist} \\
				\hline RPN \cite{ren2015faster} + \cite{girshick2015fast}  & 56.36 & 46.36 & 42.77 & N/A & N/A & N/A \\				
				\hline Our RPN + \cite{girshick2015fast}  & 71.00 & 55.88 & 51.72 & N/A & N/A & N/A \\
				\hline
				\hline \textbf{Ours} w/o Pose  & 71.12 & 57.52 & 53.77 & N/A & N/A & N/A \\				
				\hline \textbf{Ours} w/o Extra  & 71.23 & 55.56 & 51.61 & 61.89 & 47.30 & 43.69 \\
				\hline \textbf{Ours} Full  & \textbf{74.92} & \textbf{59.13} & \textbf{55.03} & \textbf{65.79} & \textbf{50.46} & \textbf{46.57} \\													
				\hline
			\end{tabular}
			\caption{Comparison of detection networks on KITTI val set.}
			\label{table:det_val_2}
			
		}\vspace{-2mm}
	}
\end{table}

\begin{table} \setlength{\tabcolsep}{1pt}
	{\tiny
		\centering{
			\begin{tabular}{|l||c|c|c||c|c|c||c|c|c|}
				\hline Methods  & Easy  & Moderate & Hard & Easy  & Moderate & Hard & Easy  & Moderate & Hard \\
				\hline & \multicolumn{3}{|c||}{Detection \& Segmentation (ASA)} & \multicolumn{3}{|c||}{Detection \& 3D Loc.$<$2m (ALP)} & \multicolumn{3}{|c|}{Detection \& 3D Loc.$<$1m (ALP)}\\
				\hline DPM \cite{felzenszwalb2010object} & 38.09  & 29.42  & 22.65 & 40.21  & 29.02  & 22.36  & 24.44  & 18.04  & 14.13 \\
				\hline 3DVP \cite{xiang2015data}  & 65.73 & 54.60 & 45.62 &  66.56 & 51.52 & 42.39 & \textbf{45.61} & \textbf{34.28} & \textbf{27.72} \\
				\hline \textbf{Ours} & \textbf{73.64}  & \textbf{66.22} & \textbf{56.34}  & \textbf{70.52} & \textbf{56.20} & \textbf{47.03} & 39.28 & 31.04 & 25.96 \\
				\hline
			\end{tabular}
			\caption{2D segmentation and 3D location of car on KITTI val set.}
			\label{table:seg_3d}
			\vspace{-2mm}
		}
	}
\end{table}

\begin{figure*} \small
	\centering
	\includegraphics[height = 0.25\linewidth,width = 0.9\linewidth]{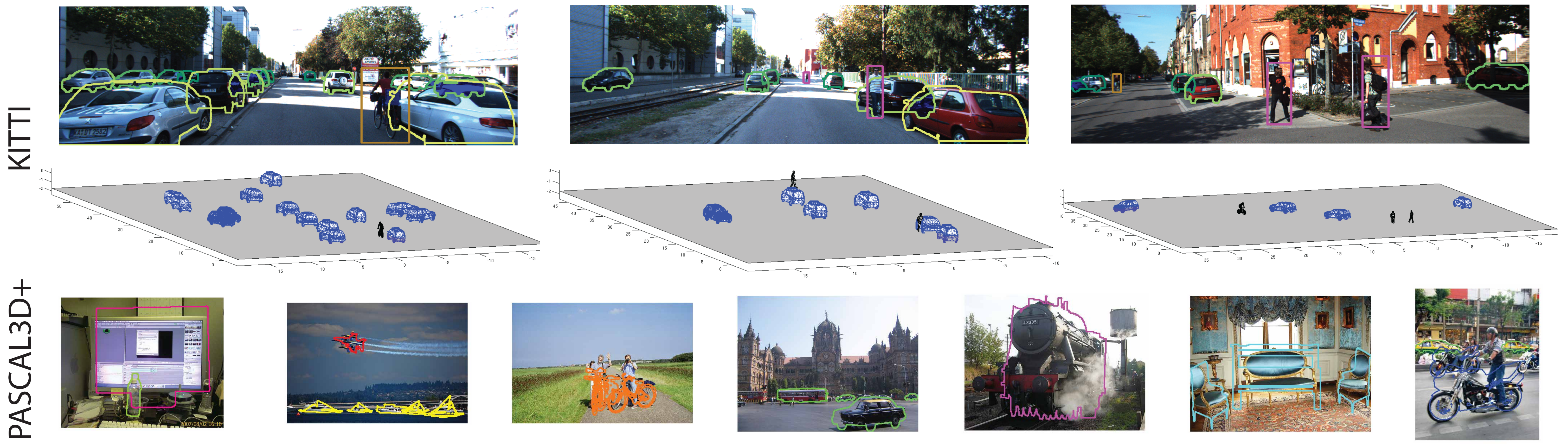}
	\caption{Examples of detections from our method. Detections with score larger than 0.5 on KITTI and 0.7 on PASCAL3D+ are shown.}
	\label{fig:results}
	\vspace{-2mm}
\end{figure*}

\begin{table}
	{\tiny
		\centering{
			\begin{tabular}{|@{\hspace{0.5em}}l@{\hspace{0.5em}}||c@{\hspace{0.5em}}|@{\hspace{0.25em}}c@{\hspace{0.25em}}|@{\hspace{0.5em}}c||c@{\hspace{0.5em}}|@{\hspace{0.25em}}c@{\hspace{0.25em}}|@{\hspace{0.5em}}c@{\hspace{0.5em}}|}
				\hline & \multicolumn{3}{|c||}{Object Detection (AP) } & \multicolumn{3}{|c|}{Orientation (AOS)}\\
				\hline Methods  & Easy  & Moderate & Hard & Easy  & Moderate & Hard \\
				\hline \multicolumn{7}{|c|}{Car} \\
				\hline ACF~\cite{DollarPAMI14pyramids} & 55.89 & 54.74 & 42.98 & N/A & N/A & N/A \\				 
				\hline DPM~\cite{felzenszwalb2010object} & 68.02 & 56.48 & 44.18 & 67.27 & 55.77 & 43.59 \\
				\hline DPM-VOC+VP~\cite{bojan15pami} & 74.95 & 64.71 & 48.76 & 72.28 & 61.84 & 46.54 \\				
				\hline OC-DPM \cite{pepikj2013occlusion} & 74.94 & 65.95 & 53.86 & 73.50 & 64.42 & 52.40 \\
				\hline SubCat \cite{subcat15} & 84.14 & 75.46 & 59.71 & 83.41 & 74.42 & 58.83 \\								
				\hline Regionlets \cite{wang2013regionlets} & 84.75 & 76.45 & 59.70 & N/A & N/A & N/A\\
				\hline AOG \cite{li2014integrating} & 84.80 & 75.94 & 60.70 & 33.79 & 30.77 & 24.75 \\
				\hline Faster R-CNN \cite{ren2015faster} & 86.71 & 81.84 & 71.12 & N/A & N/A & N/A \\						
				\hline 3DVP \cite{xiang2015data} & 87.46 & 75.77 & 65.38 & 86.92 & 74.59 & 64.11\\
				\hline 3DOP \cite{3DOPNIPS15}  & \textbf{93.04} & 88.64 & 79.10 & \textbf{91.44} & 86.10 & 76.52 \\
				\hline Mono3D \cite{XiaozhiCVPR16} & 92.33 & 88.66 & 78.96 & 91.01 & 86.62 & 76.84 \\
				\hline SDP+RPN \cite{yang2016sdp} & 90.14 & 88.85 & 78.38 & N/A & N/A & N/A \\
				\hline MS-CNN \cite{MSCNN2016} & 90.03 & 89.02 & 76.11 & N/A & N/A & N/A \\
				\hline SubCNN-VGG16 (\textbf{Ours}) & 90.74 & 88.55 & 77.95 & 90.49 & 87.88 & 77.10 \\
				\hline SubCNN-GoogleNet (\textbf{Ours}) & 90.81 & \textbf{89.04} & \textbf{79.27} & 90.67 & \textbf{88.62} & \textbf{78.68} \\
				
				\hline \multicolumn{7}{|c|}{Pedestrian} \\
				\hline ACF~\cite{DollarPAMI14pyramids} & 44.49 & 39.81 & 37.21  & N/A & N/A & N/A \\					
				\hline DPM~\cite{felzenszwalb2010object} & 47.74 & 39.36 & 35.95 & 43.58 &	35.49 &	32.42 \\
				\hline DPM-VOC+VP~\cite{bojan15pami} & 59.48 &	44.86 &	40.37 & 53.55 &	39.83 &	35.73  \\				
				\hline FilteredICF \cite{Zhang2015FilteredICF} & 67.65 & 56.75 & 51.12 & N/A & N/A & N/A\\
				\hline DeepParts \cite{DeepParts2015} & 70.49 & 58.67 & 52.78 & N/A & N/A & N/A\\
				\hline Regionlets \cite{wang2013regionlets} & 73.14 & 61.15 & 55.21 & N/A & N/A & N/A\\
				\hline Faster R-CNN \cite{ren2015faster} & 78.86 & 65.90 & 61.18 & N/A & N/A & N/A \\					
				\hline Mono3D \cite{XiaozhiCVPR16} & 80.35 & 66.68 & 63.44 & 71.15 & 58.15 & 54.94 \\ 	
				\hline 3DOP \cite{3DOPNIPS15}  & 81.78 & 67.47 & 64.70 & 72.94 & 59.80 & 57.03 \\
				\hline SDP+RPN \cite{yang2016sdp} & 80.09 & 70.16 & 64.82 & N/A & N/A & N/A \\				
				\hline MS-CNN \cite{MSCNN2016} & \textbf{83.92} & \textbf{73.70} & \textbf{68.31} & N/A & N/A & N/A \\				
				\hline SubCNN-VGG16 (\textbf{Ours}) & 79.13 & 66.13 & 61.27 & 72.61 & 59.40 & 54.78 \\
				\hline SubCNN-GoogleNet (\textbf{Ours}) & 83.28 & 71.33 & 66.36 & \textbf{78.45} & \textbf{66.28} & \textbf{61.36} \\
				
				\hline \multicolumn{7}{|c|}{Cyclist} \\
				\hline DPM~\cite{felzenszwalb2010object} & 35.04 &	27.50 &	26.21 & 27.54 &	22.07 &	21.45  \\
				\hline DPM-VOC+VP~\cite{bojan15pami} & 42.43 &	31.08 &	28.23  & 30.52 & 23.17 & 21.58 \\
				\hline Regionlets \cite{wang2013regionlets} & 70.41 & 58.72 & 51.83  & N/A & N/A & N/A \\
				\hline Faster R-CNN \cite{ren2015faster} & 72.26 & 63.35 & 55.90 & N/A & N/A & N/A \\
				\hline Mono3D \cite{XiaozhiCVPR16} & 76.04 & 66.36 & 58.87 & 65.56 & 54.97 & 48.77 \\ 								
				\hline 3DOP \cite{3DOPNIPS15}  & 78.39 & 68.94 & 61.37 & 70.13 & 58.68 & 52.35 \\
				\hline SDP+RPN \cite{yang2016sdp} & 81.37 & 73.74 & 65.31 & N/A & N/A & N/A \\				
				\hline MS-CNN \cite{MSCNN2016} & \textbf{84.06} & \textbf{75.46} & \textbf{66.07} & N/A & N/A & N/A \\				
				\hline SubCNN-VGG16 (\textbf{Ours}) & 74.40 & 61.98 & 54.75  & 63.74 & 52.06 & 45.93 \\
				\hline SubCNN-GoogleNet (\textbf{Ours}) & 79.48 & 71.06 & 62.68  & \textbf{72.00} & \textbf{63.65} & \textbf{56.32} \\
				\hline
			\end{tabular}
			\caption{Comparison between different methods on KITTI test set.}
			\label{table:test}
			\vspace{-4mm}
		}
	}
\end{table}

In Table \ref{table:det_val}, by using region proposals from our RPN, we achieve better performance on detection and orientation estimation across all the three categories. The experimental results demonstrate the advantages of our RPN. We also tried to share the conv layers in our RPN and our detection network. However, since the architecture of our RPN after the conv layers for feature extraction is quite different from that of the detection network, we found that the training cannot converge, which verifies our observation that the RPN and the detection network have developed their own conv features that are suitable for its own task.

\noindent \textbf{Detection Network Evalutaion.} In Table \ref{table:det_val_2}, we first show that our RPN achieves significantly better performance than the RPN in \cite{ren2015faster} when the two RPNs are used with Fast R-CNN \cite{girshick2015fast} on the KITTI validation set respectively. Then, we use region proposals from our RPN and compare different variations of the network architecture for detection. i) ``Ours w/o Pose'' indicates using 2D subcategories from clustering on 2D appearances of objects without using additional pose information. As we can see, our method still outperforms Fatser R-CNN \cite{ren2015faster} in this case. ii) By using pose information to obtain subcategories, our detection network is also able to estimate the orientation of the object. ``Ours w/o Extra'' refers to a network without feature extrapolating. By augmenting the network with the feature extrapolating layer, our full model (``Ours Full'' in Table \ref{table:det_val_2}) further boosts the performance, except for a minor drop on orientation estimation of pedestrian.

\begin{table*}\setlength{\tabcolsep}{5pt}
	\tiny
	\centering
	\begin{tabular}{|l||c|c|c|c|c|c|c|c|c|c|c|c||c|}
		\hline
		Methods & aeroplane & bicycle & boat & bottle & bus & car & chair & diningtable & motorbike & sofa & train & tvmonitor & Average\\
		\hline
		\hline \multicolumn{14}{|c|}{Object Detection (AP)} \\
		\hline DPM \cite{felzenszwalb2010object} & 42.2 & 49.6 & 6.0 & 20.0 & 54.1 & 38.3 & 15.0 & 9.0 & 33.1 & 18.9 & 36.4 & 33.2 & 29.6 \\
		\hline R-CNN \cite{girshick2013rich} & 72.4 & 68.7 & 34.0 & -- & 73.0 & 62.3 & 33.0 & 35.2 & 70.7 & 49.6 & 70.1 & 57.2 & 56.9 \\
		\hline \textbf{Ours} w/o Extra & 76.3 & 73.4 & \textbf{43.4} & 44.7 & \textbf{74.5} & 63.3 & 35.4 & 32.4 & 74.9 & 51.9 & 74.1 & 60.9 & 58.8 \\
		\hline \textbf{Ours} Full & \textbf{76.5} & \textbf{74.0} & 42.4 & \textbf{47.0} & \textbf{74.5} & \textbf{64.7} & \textbf{38.5} & \textbf{38.6} & \textbf{76.7} & \textbf{55.1} &  \textbf{74.8} & \textbf{65.3} & \textbf{60.7} \\
		\hline
		\hline \multicolumn{14}{|c|}{Joint Object Detection and Pose Estimation (4 Views AVP)} \\
		\hline VDPM \cite{xiang2014beyond} & 34.6 & 41.7 & 1.5 & -- & 26.1 & 20.2 & 6.8 & 3.1 & 30.4 & 5.1 & 10.7 & 34.7 & 19.5 \\		
		\hline DPM-VOC+VP \cite{bojan15pami} & 39.4 & 43.9 & 0.3 & -- & 49.1 & 37.6 & 6.1 & 3.0 & 32.2 & 11.8 & 12.5 & 33.2 & 24.5 \\
		\hline \textbf{Ours} w/o Extra & \textbf{62.3} & 56.6 & 18.0 & -- & 62.0 & 40.9 & 19.3 & 14.9 & \textbf{62.3} & 44.1 & \textbf{58.1} & 58.5 & 45.2  \\
		\hline \textbf{Ours} Full & 61.4 & \textbf{60.4} & \textbf{21.1} & -- & \textbf{63.0} & \textbf{48.7} & \textbf{23.8} & \textbf{17.4} & 60.7 & \textbf{47.8} & 55.9 & \textbf{62.3} & \textbf{47.5} \\
		\hline
		\hline \multicolumn{14}{|c|}{Joint Object Detection and Pose Estimation (8 Views AVP)} \\
		\hline VDPM \cite{xiang2014beyond} & 23.4 & 36.5 & 1.0 & -- & 35.5 & 23.5 & 5.8 & 3.6 & 25.1 & 12.5 & 10.9 & 27.4 & 18.7 \\		
		\hline DPM-VOC+VP \cite{bojan15pami} & 29.7 & \textbf{42.6} & 0.4 & -- & 39.5 & 36.8 & 9.4 & 2.6 & 32.9 & 11.0 & 10.3 & \textbf{28.6} & 22.2 \\
		\hline \textbf{Ours} w/o Extra & 45.9 & 25.5 & 11.1 & -- & 37.7 & 34.6 & 15.2 & 7.4 & \textbf{37.1} & \textbf{33.0} & 42.5 & 24.3 & 28.6 \\
		\hline \textbf{Ours} Full & \textbf{48.8} & 36.3 & \textbf{16.4} & -- & \textbf{39.8} & \textbf{37.2} & \textbf{19.1} & \textbf{13.2} & 37.0 & 32.1 & \textbf{44.4} & 26.9 & \textbf{31.9} \\	
		\hline	
		\hline \multicolumn{14}{|c|}{Joint Object Detection and Pose Estimation (16 Views AVP)} \\
		\hline VDPM \cite{xiang2014beyond} & 15.4 & 18.4 & 0.5 & -- & 46.9 & 18.1 & 6.0 & 2.2 & 16.1 & 10.0 & 22.1 & 16.3 & 15.6 \\		
		\hline DPM-VOC+VP \cite{bojan15pami} & 17.0 & \textbf{24.7} & 1.0 & -- & 49.0 & 30.1 & 6.6 & 3.0 & 17.2 & 7.7 & 20.4 & 20.2 & 17.9 \\
		\hline \textbf{Ours} w/o Extra & 23.3 & 19.2 & 8.4 & -- & \textbf{52.6} & 27.0 & 9.9 & 5.1 & \textbf{23.6} & \textbf{20.9} & 27.4 & \textbf{27.9} & 22.3 \\
		\hline \textbf{Ours} Full & \textbf{28.0} & 23.7 & \textbf{10.7} & -- & 50.8 & \textbf{31.4} & \textbf{14.3} & \textbf{9.4} & 23.4 & 19.5 & \textbf{30.7} & 27.8 & \textbf{24.5} \\					
		\hline
		\hline \multicolumn{14}{|c|}{Joint Object Detection and Pose Estimation (24 Views AVP)} \\
		\hline VDPM \cite{xiang2014beyond} & 8.0 & 14.3 & 0.3 & -- & 39.2 & 13.7 & 4.4 & 3.6 & 10.1 & 8.2 & 20.0 & 11.2 & 12.1 \\		
		\hline DPM-VOC+VP \cite{bojan15pami} & 10.6 & \textbf{16.7} & 2.2 & -- & \textbf{43.5} & \textbf{25.4} & 4.4 & 2.3 & 11.3 & 4.9 & 22.4 & 14.4 & 14.4 \\
		\hline \textbf{Ours} w/o Extra & 18.9 & 10.5 & 6.7 & -- & 34.3 & 23.3 & 8.3 & 6.5 & \textbf{20.6} & 17.5 & \textbf{33.8} & 17.0 & 17.9 \\
		\hline \textbf{Ours} Full & \textbf{20.7} & 16.4 & \textbf{7.9} & -- & 34.6 & 24.6 & \textbf{9.4} & \textbf{7.6} & 19.9 & \textbf{20.0} & 32.7 & \textbf{18.2} & \textbf{19.3} \\						
		\hline		
		
	\end{tabular}
	\caption{AP/AVP Comparison between different methods on the PASCAL3D+ dataset.}
	\label{tab:pascal3d}
	\vspace{-2mm}
\end{table*}

\begin{table*} \setlength{\tabcolsep}{4pt}
	\tiny
	\centering
	\begin{tabular}{l|c|c c c c c c c c c c c c c c c c c c c c} 
		\hline
		& mAP & aero & bike & bird & boat & bottle & bus & car & cat & chair & cow & table & dog & horse & mbike & person & plant & sheep & sofa & train & tv \\
		\hline
		\cite{girshick2015fast} & 66.9 & \textbf{74.5} & 78.3 & 69.2 & 53.2 & 36.6 & 77.3 & 78.2 & 82.0 & 40.7 & 72.7 & \textbf{67.9} & 79.6 & 79.2 & 73.0 & 69.0 & 30.1 & 65.4 & \textbf{70.2} & 75.8 & 65.8 \\
		\cite{ren2015faster} & \textbf{69.9} & 70.0 & \textbf{80.6} & \textbf{70.1} & 57.3 & \textbf{49.9} & 78.2 & \textbf{80.4} & 82.0 & \textbf{52.2} & \textbf{75.3} & 67.2 & 80.3 & 79.8 & 75.0 & \textbf{76.3} & \textbf{39.1} & \textbf{68.3} & 67.3 & \textbf{81.1} & \textbf{67.6} \\
		Ours & 68.5 & 70.2 & 80.5 & 69.5 & \textbf{60.3} & 47.0 & \textbf{79.0} & 78.7 & \textbf{84.2} & 48.5 & 73.9 & 63.0 & \textbf{82.7} & \textbf{80.6} & \textbf{76.0} & 70.2 & 38.2 & 62.4 & 67.7 & 77.7 & 60.5 \\	
		\hline		
		
	\end{tabular}
	\caption{AP comparison between Fast R-CNN \cite{girshick2015fast}, Faster R-CNN \cite{ren2015faster} and our method on PASCAL VOC 2007 dataset.}
	\label{tab:pascal2007}
	\vspace{-4mm}
\end{table*}

\noindent \textbf{Evaluation on 2D Segmentation and 3D Localization.} 3DVPs enable us to transfer the meta data to the detect objects, so our method is able to segment the boundary of object. In addition, after detecting the objects and estimating their 3D poses, we can back-project them into 3D using the camera parameters provided in KITTI, so as to evaluate the 3D localization performance. In table \ref{table:seg_3d}, we compare our method on 2D segmentation and 3D localization of car with DPM \cite{felzenszwalb2010object} and 3DVP \cite{xiang2015data} on the KITTI validation set. We have significantly improve the segmentation accuracy and 3D location accuracy when the 2-meter threshold is used (i.e., a detection within  2 meters from the ground truth location is considered to be correct). Surprisingly, \cite{xiang2015data} obtains better 3D localization accuracy with the 1-meter threshold, which indicates that more detections from \cite{xiang2015data} are within the 1-meter distance from the ground truth.

\subsection{KITTI Test Set Evaluation}

To compare with the state-of-the-art methods on the KITTI detection benchmark, we train our RPN and detection network with all the KITTI training data, and then test our method on the KITTI test set by submitting our results to \cite{KITTIObject}. Table \ref{table:test} presents the detection and orientation estimation results on the three categories, where we compare our method (SubCNN) with different methods evaluated on KITTI. We have experimented fine-tuning both the VGG16 network and the GoogleNet for the detection network. Our method ranks on top among all the published methods. The experimental results demonstrate the ability of our CNNs in using subcategory information for detection and orientation estimation. Fig. \ref{fig:results} presents some examples of our detection and 3D localization results on KITTI.

\subsection{Evaluation on PASCAL3D+ and PASCAL VOC}

We also evaluate our detection framework on the 12 categories in PASCAL3D+. Table \ref{tab:pascal3d} presents the detection results in AP and the joint detection and pose estimation results in AVP. After generating region proposals from our RPN, we experiment with our detection networks with and without feature extrapolation. First, in terms of detection, our method improves over R-CNN \cite{girshick2013rich} on all 12 categories. Second, in terms of join detection and pose estimation, our method significantly outperforms two state-of-the-art methods: VDPM \cite{xiang2014beyond} and DPM-VOC+VP \cite{bojan15pami}. Third, feature extrapolation helps both detection and pose estimation on PASCAL3D+. It is worth mentioning that PASCAL3D+ has much fewer training examples in each subcategory compared to KITTI. Our pose estimation performance is limited by the number of training examples available in PASCAL3D+. We also note that the two recent methods \cite{tulsiani2014viewpoints,su2015render} achieve very appealing pose estimation results on PASCAL3D+. However, both of them utilize additional training images (ImageNet images in \cite{tulsiani2014viewpoints} and synthetic images in \cite{su2015render}) and conduct detection and pose estimation with separate CNNs, where a CNN is specifically designed for pose estimation. Our method is capable of simultaneous object detection and viewpoint estimation even in the presence of limited training examples per viewpoint. Fig. \ref{fig:results} shows some detection results from our method. We again transfer segmentation masks of 3DVPs to the detected objects according to the subcategory classification results. Please see supplementary material for more examples.

To demonstrate that our method also works on datasets with bounding box annotations only, we have conducted experiments on the PASCAL VOC 2007 dataset, where subcategories are obtained by clustering on image features. In table \ref{tab:pascal2007}, we compare with Fast R-CNN \cite{girshick2015fast} and Faster R-CNN \cite{ren2015faster}. We have achieved comparable performance to the state-of-the-arts. Region proposal on PASCAL VOC is relatively easy compared to KITTI. So we do not see much improvement with our RPN on PASCAL VOC 2007.

\section{Conclusion}

In this work, we explore how subcategory information can be exploited in CNN-based object detection. We have proposed a novel region proposal network, and a novel object detection network, where we explicitly employ subcategory information to improve region proposal generation, object detection and object pose estimation. Our subcategory-aware CNNs can also handle the scale variation of objects using image pyramids in an efficient way. We have conducted extensive experiments on the KITTI detection benchmark, the PASCAL3D+ dataset and PASCAL VOC 2007 dataset. Our method achieves the state-of-the-art results on these benchmarks.

\noindent \textbf{Acknowledgments.} We acknowledge the support of Nissan grant 1188371-1-UDARQ and MURI grant 1186514-1-TBCJE.

{\small
	\bibliographystyle{ieee}
	\bibliography{egbib}
}
\end{document}